\documentclass[runningheads]{llncs}

\usepackage{eccv}

\usepackage{eccvabbrv}
\usepackage{makecell}
\usepackage{graphicx}
\usepackage{booktabs}
\usepackage{wrapfig}

\usepackage[accsupp]{axessibility}  

\usepackage{hyperref}

\usepackage{orcidlink}

\begin{document}

\title{TryOnCrafter: Unleashing Camera Trajectories for Realistic Video Virtual Try-on via a Renderable 4D Try-on Proxy} 

\titlerunning{TryOnCrafter}

\author{Hao Sun\inst{1,2,3} \and
Hao Yan\inst{2} \and
Mengting Chen\inst{2}\textsuperscript{\textdagger}\textsuperscript{*} \and
Quanjian Song\inst{2,4} \and
Yu Li\inst{1} \and
Juan Cao\inst{1} \and
Jinsong Lan\inst{2} \and
Xiaoyong Zhu\inst{2} \and
Bo Zheng\inst{2} \and
Sheng Tang\inst{1}\textsuperscript{*}}

\authorrunning{Sun et al.}

\institute{
Institute of Computing Technology, Chinese Academy of Sciences \\
\and Taobao \& Tmall Group of Alibaba \\
\and 
University of Chinese Academy of Sciences\\
\and Xiamen University \\[0.5em]
Project Page: \url{https://sunhao242.github.io/TryOnCrafter_web.github.io/}
}

\maketitle

\begingroup
\renewcommand{\thefootnote}{}
\footnotetext{
\textsuperscript{\textdagger} Project leader\\
\textsuperscript{*} Co-corresponding authors.
}
\endgroup

\begin{figure*}
    \centering
    \setlength{\belowcaptionskip}{-0.8cm}
\includegraphics[width=1.0\linewidth]{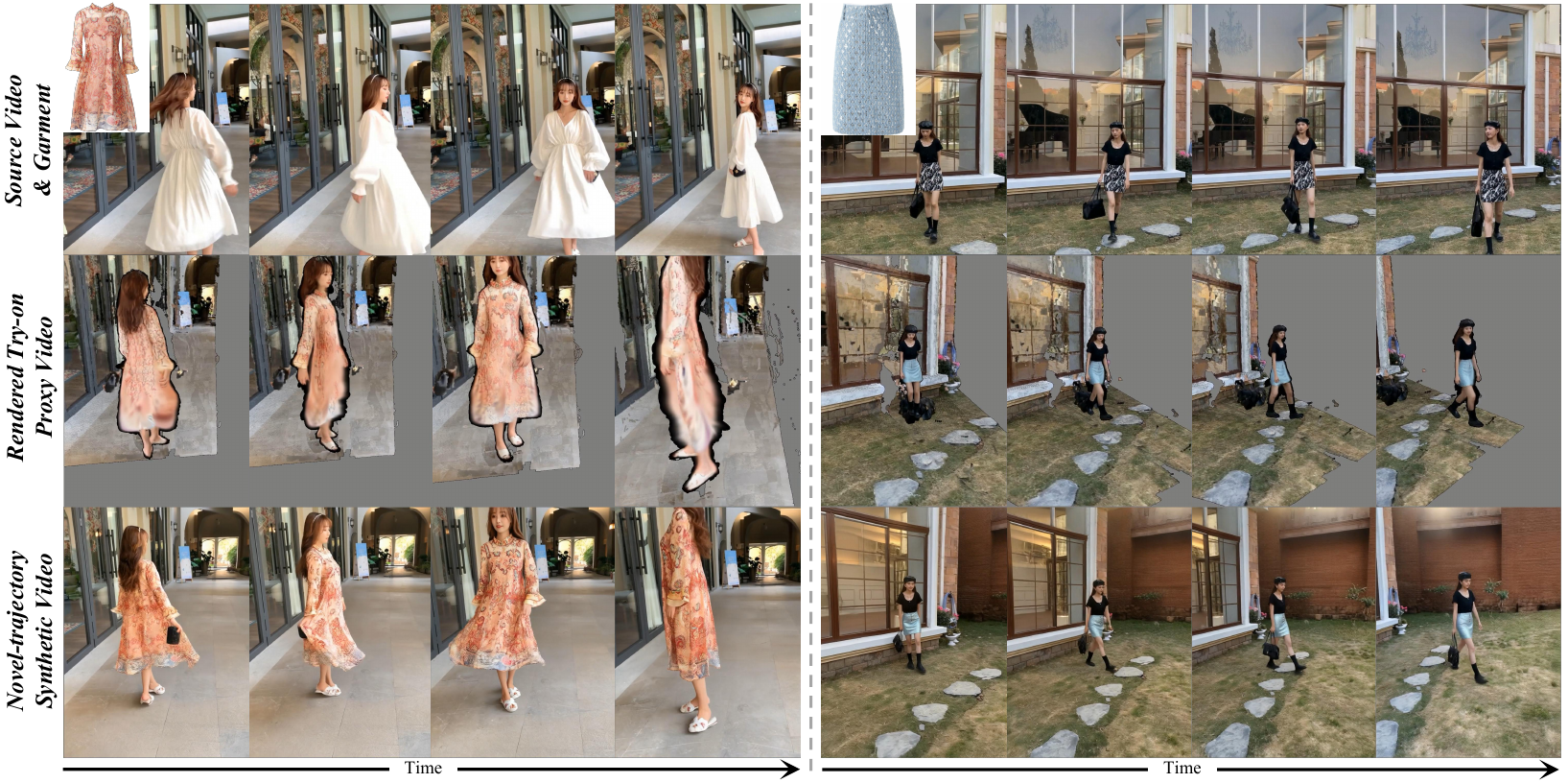}
  \caption{Examples synthesized by TryOnCrafter.  We introduce a Renderable 4D Try-on Proxy (middle) as a geometric anchor to guide the Video Diffusion Transformer. This explicit 4D representation enables photorealistic try-on across unconstrained, novel camera trajectories (bottom) not present in the source video (top). }
  \label{fig:teaser}
\end{figure*}

\begin{abstract}
While Video Virtual Try-on (VVT) has achieved remarkable progress in synthesizing realistic garment overlays on dynamic subjects, existing paradigms remains fundamentally constrained by a passive dependency on source camera trajectories, failing to accommodate the requisite interactive freedom for omnidirectional viewpoint exploration. 
To address this limitation, we define a pioneering research frontier: Camera-controllable Video Virtual Try-on (CaM-VVT). Unlike conventional VVT, CaM-VVT not only necessitates viewpoint-agnostic texture hallucination but also strict structural synchronization between non-rigid human dynamics and background contexts under arbitrary, unconstrained camera movements.
To tackle these challenges, we present TryOnCrafter, the first unified DiT-based framework specifically architected for the CaM-VVT task. Departing from implicit pixel-space manipulation, we introduce a Renderable 4D Try-on Proxy that explicitly decouples the human subject from the environment. This is achieved by distilling high-fidelity 2D try-on priors into a clothed 3DGS-based avatar, which is subsequently animated via SMPL-X sequences and metric-aligned into a reconstructed background point cloud. This proxy establishes a robust structural foundation with superior texture density and motion integrity. Our Proxy-Anchored Video DiT leverages this robust structural foundation as a primary geometric anchor, ensuring that the synthesized photorealistic videos are strictly constrained by prescribed trajectories and physically plausible deformations. 
Benefiting from the inherent editability of the 4D proxy, TryOnCrafter facilitates diverse downstream applications, including human relocalization, ``bullet time'' effects, and $360$-degree orbital viewing. 
Extensive experiments on our established CaM-VVTBench demonstrate that TryOnCrafter significantly outperforms existing baselines in preserving structural consistency and garment identity across complex camera maneuvers.
\keywords{Video Virtual Try-on \and Camera Control \and 4D Reconstruction}
\end{abstract}
\section{Introduction}
Video Virtual Try-on (VVT) aims to overlay target garments onto dynamic human subjects videos while preserving spatio-temporal consistency.
As a critical engine for immersive ``try-before-you-buy'' experiences, VVT has become a cornerstone of digital fashion and e-commerce, increasingly serving as a focal point for research.
Early studies predominantly rely on image-based warping or frame-wise priors~\cite{xu2024tunnel, karras2024fashion, wang2024gpd}.
To further enhance generative stability in intricate environments, recent works~\cite{li2025magictryon, zuo2025dreamvvt} integrate powerful DiT-based architectures, pushing the boundaries of VVT toward more diverse and complex scenarios.

Despite this progress, existing VVT research shares a fundamental limitation: \textbf{they are confined to the fixed camera trajectory of the input monocular video}.
Such limitation conflicts with real try-on needs, where users expect to freely rotate and inspect garments from novel viewpoints to assess details such as side profiles and back appearance.
In light of that, this paper introduce a new frontier: Camera-controllable Video Virtual Try-On (\textbf{CaM-VVT}).
By decoupling synthesis from the input trajectory, CaM-VVT enables interactive, 3D-aware virtual fitting beyond passive video replay.
It requires view-consistent garment synthesis from novel viewpoints while maintaining strict geometric alignment with the background under arbitrary camera motion.

A straightforward approach for CaM-VVT is a two-stage pipeline: first perform video virtual try-on, then apply a video-to-video (V2V) camera control model.
However, such sequential design remains suboptimal for digital fashion due to three fundamental hurdles:
(i) \textit{Cascaded error accumulation.} Texture inconsistencies from the initial video virtual try-on stage are often amplified by the subsequent V2V camera control models. This is because the control models may not handle the out-of-distribution try-on results.
(ii) \textit{Limited garment modeling.} Existing V2V camera control methods lack explicit modeling of human geometry and garment deformation. As a result, directly applying them to try-on videos often fails to preserve the spatio-temporal coherence of garment structure.
(iii) \textit{Heavy computational burden.} Both current video virtual try-on and V2V camera control models are heavy and slow at inference. Naively cascading them makes end-to-end inference impractical and can nearly double the runtime.

To overcome these hurdles, we present \textbf{TryOnCrafter}, a unified and end-to-end Diffusion Transformer (DiT) framework for CaM-VVT.
In detail, TryOnCrafter consists of two core components:
(i) \textbf{4D Try-on Proxy.}
Unlike prior camera-control works~\cite{yu2024viewcrafter,yu2025trajectorycrafter,xiao2024trajectory,wu2025cat4d,liu2024novel} that rely on and often fragmented dynamic point cloud to represent the entire 4D scene, our proxy explicitly decouples the human subject from the environment.
By distilling high-quality 2D image try-on priors into a clothed 3DGS-based avatar and synchronizing it with SMPL-X motion sequences, we achieve a complete reconstruction of both the human geometry and its articulated dynamics. Compared to sparse and unstable point cloud, this 4D try-on proxy provides significantly more robust and comprehensive garment textures and motion cues, ensuring structural integrity even under radical viewpoint transitions.
(ii) \textbf{Proxy-Anchored Try-on Video Diffusion Transformer.}
Building upon this foundation, we introduce the Proxy-Anchored Video Diffusion Transformer. This model leverages the rendered proxy sequence as a primary structural anchor to synthesize photorealistic try-on videos. By grounding the generative process in explicit 4D guidance, TryOnCrafter ensures that the synthesized results are not only visually stunning but also strictly constrained by physically plausible deformations and target camera trajectories.

To further support our task, we establish \textbf{CaM-VVTBench}, a comprehensive evaluation protocol for controllable try-on.
Extensive experiments show TryOnCrafter significantly outperforms existing baselines in preserving structural consistency and garment identity.
Moreover, TryOnCrafter supports downstream applications, including human relocalization, ``bullet time'' effects and  $360^{\circ}$ orbital viewing, paving a new way for interactive digital fashion experiences.

In summary, the contributions of this paper are threefold:
\begin{itemize}
    \item We propose TryOnCrafter, the first framework to extend video virtual try-on to the Camera-controllable Video Virtual Try-on (CaM-VVT) task, decoupling synthesis from the constraints of the input camera trajectory.
    \item We introduce the Renderable 4D Try-on Proxy combined with a Proxy-Anchored Video DiT, establishing a pioneering re-rendering-based paradigm that provides robust structural grounding and motion integrity.
    \item We establish CaM-VVTBench, a comprehensive evaluation protocol for controllable try-on. Extensive experiments demonstrate that TryOnCrafter significantly outperforms existing baselines in preserving structural consistency and garment identity under unconstrained camera movements.
\end{itemize}

\section{Related Work}

\noindent\textbf{Video Virtual Try-On.}
Recent advances in video diffusion models have substantially advanced video virtual try-on~\cite{xu2024tunnel,nguyen2025swifttry,karras2024fashion,he2024wildvidfit,wang2024gpd,guo2025any2anytryon,song2026fashionchameleon}, thereby meeting the e-commerce demand for dynamic visualization. This task aims to transfer a target garment onto a person in a source video while preserving spatio-temporal consistency.
Early research primarily relied on image diffusion models to generate videos. For example, ViViD~\cite{fang2024vivid} enhances temporal coherence by embedding temporal attention mechanisms and additionally constructs a large-scale VVT dataset to facilitate benchmarking. With the advancement of diffusion transformer architectures, subsequent studies aim at more scalable and expressive generation. CatV$^2$TON~\cite{chong2025catv2ton} directly concatenates garment and video latents for multimodal attention interaction, eliminating the reliance on extra auxiliary networks. Magic-Tryon~\cite{li2025magictryon} develops a refined feature injection strategy to maintain detailed garment textures and logos. More recently, DreamVVT~\cite{zuo2025dreamvvt} extends video try-on task from simple indoor scenes to complex outdoor environments. Despite these advancements, existing methods lack explicit camera modeling, restricting the output to fixed viewpoints. This inherently falls short of the growing virtual fitting demand for dynamic and multi-view try-on experiences. To fill this gap, this work pioneers re-rendering-based paradigm, unifying video try-on and camera-controllable video generation into a single framework.

\noindent\textbf{Camera Control in Video-to-Video Generation.}
Driven by industry demand, camera-controllable video generation has become a focal point in controllable diffusion models~\cite{huang2025arteditor,li2024latentsync,zhang2025easycontrol,song2025worldwander,lin2024pair,zhang2025stable,song2025univst,gong2025relationadapter,song2025scenedecorator,song2025makeanything,li2024director3d}.
Pioneering methods for camera-controllable video generation~\cite{he2024cameractrl,wang2024motionctrl,cao2025uni3c} typically rely on specific external conditions. To improve efficiency, subsequent approaches~\cite{hou2024training, song2025lightmotion, meng2024nvs} have transitioned toward simulating camera motion through a tuning-free paradigm.
Recently, some works~\cite{bai2025recammaster,yu2024viewcrafter,liu2024novel,xiao2024trajectory,yu2025trajectorycrafter,wu2025cat4d} have integrated camera control into video-to-video translation to achieve perspective redirection, which is most relevant with our task.
ReCamMaster~\cite{bai2025recammaster} injects extrinsic embeddings to enable perspective translation, but the absence of explicit geometry often leads to physical inconsistencies. While TrajectoryCrafter~\cite{yu2025trajectorycrafter} and CAT4D~\cite{wu2025cat4d} explicitly reconstruct dynamic 3D scenes using point clouds to ensure physical consistency, it fails to decouple camera motion from human motion. In contrast, our TryOnCrafter is the first to unify virtual try-on, human motion, and camera trajectory within a single video generation framework, which is unexplored in prior works.


\section{Method}
\begin{figure*}[t]
    \centering
    \setlength{\belowcaptionskip}{-0.5cm}
\includegraphics[width=1.0\linewidth]{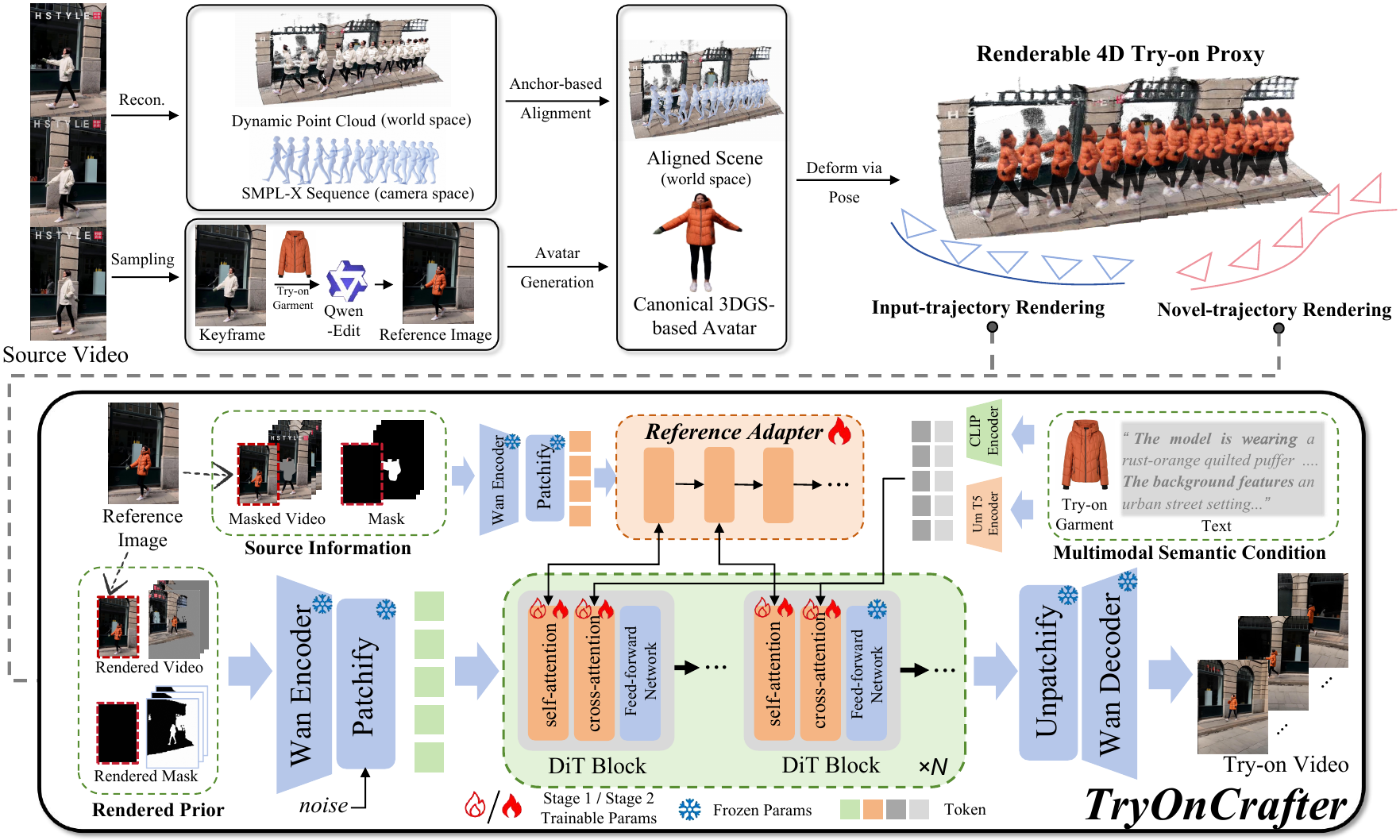}
  \caption{Overview of TryOnCrafter. \textit{Top}: 4D Try-on Proxy Construction. A metric-aligned scene in world space is established via anchor-based alignment of dynamic point clouds and SMPL-X sequences. Integrating a canonical 3DGS-based avatar distilled from reference image, the resulting 4D Try-on Proxy supports interactive editing and rendering across arbitrary camera trajectories.
\textit{Bottom}: Proxy-Anchored Try-on Video Generation. The try-on video is synthesized by the DiT under the integrated guidance of rendered priors, cross-view source features, and multimodal semantic conditions. }
  \label{fig:pipeline}
\end{figure*}

The objective of TryOnCrafter is to synthesize high-fidelity virtual try-on videos across arbitrary camera trajectories, ensuring the preservation of intricate garment textures and robust temporal coherence. As illustrated in Fig.~\ref{fig:pipeline}, our framework comprises two primary stages:  4D Try-on Proxy Construction (Sec.~\ref{subsection: 4D Try-on Proxy Construction}) and Proxy-Anchored Video Generation (Sec.~\ref{subsection: Proxy-Anchored Try-on Video Generation}).

\subsection{4D Try-on Proxy Construction.}
\label{subsection: 4D Try-on Proxy Construction}

\subsubsection{Scene Reconstruction and Anchor-based Alignment.}
Given the source video $\mathcal{V}^s$, we first reconstruct the dynamic scene from $\mathcal{V}^s$ by estimating dense depth $\mathcal{D}_t$, confidence $\mathcal{W}_t$, and camera parameters $\{\mathbf{K}_t, \mathbf{R}_t, \mathbf{t}_t\}$ via feed-forward MVS frameworks \cite{wang2025vggt,wang2025pi,zhang2024monst3r,wang2024dust3r}. The resulting back-projected global point cloud $\mathcal{P}$ is partitioned into a human component $\mathcal{P}^h$ and a background component $\mathcal{P}^b$ using SAM 2\cite{ravi2024sam}. Specifically, $\mathcal{P}^h$ captures the human geometry clothed in the source garment, it is subsequently replaced by our high-fidelity, 3DGS-based avatar dressed in the target try-on apparel. Concurrently, $\mathcal{P}^b$ is preserved as the background context to enable seamless rendering within a unified, metric-aligned world space $\mathcal{S}^w$.

\begin{figure*}
    \centering
    \setlength{\belowcaptionskip}{-0.5cm}
\includegraphics[width=1.0\linewidth]{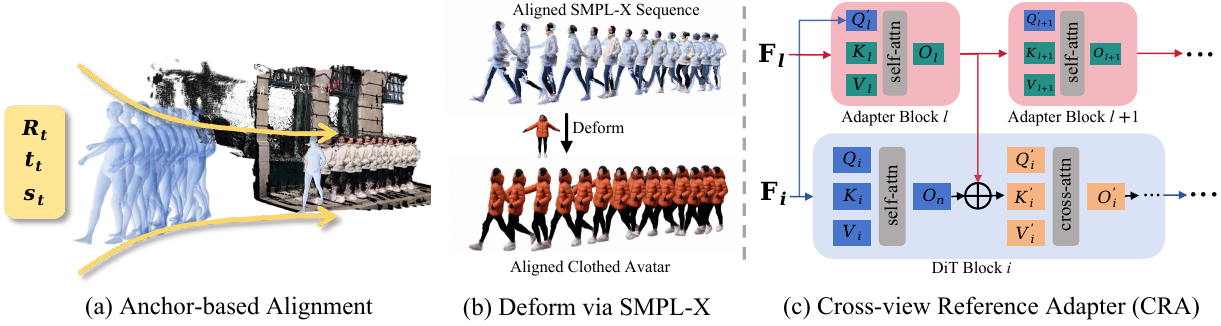}
  \caption{Details of 4D Try-on Proxy Construction and CRA Module. (a) A similarity transformation $\{R_t, t_t, s_t\}$ maps the SMPL-X sequence from camera space to the world space, utilizing the human point cloud as a geometric anchor. (b) The canonical 3DGS-based avatar is dynamically warped via LBS driven by the aligned SMPL-X sequence, preserving structural integrity across the motion. (c) The CRA module facilitates interaction between DiT features $\mathbf{F}_i$ and reference features $\mathbf{F}_l$. By sharing $K$-$V$ projections and utilizing independent $Q'$-$O'$ weights, it integrates a reference residual $\mathbf{O}_l'$ into the backbone to ensure cross-view identity and background consistency.}
  \label{fig:pipeline2}
\end{figure*}

Then, we estimate a temporally consistent SMPL-X sequence $\{\Theta_t\}_{t=1}^T$ using GVHMR~\cite{shen2024world} to decouple the human motion from $\mathcal{V}^s$. A fundamental challenge in our pipeline is the coordinate discrepancy between the parametric SMPL-X sequence $\{\Theta_t\}_{t=1}^T$, estimated in the local camera space $\mathcal{S}^c$, and the dynamic point cloud $\{\mathcal{P}_t\}_{t=1}^T$, reconstructed in the global world space $\mathcal{S}^w$. While these two representations are inherently consistent in their 2D projections, they reside in different metric scales and 3D reference frames. To bridge this gap, as illustrated in Fig.~\ref{fig:pipeline2}(a), we utilize $\{\mathcal{P}_t^h\}$ as a geometric anchor and formulate a robust similarity transformation $\mathcal{T}(\mathbf{v}) = s_tR_t\mathbf{v} + t_t$ to map the SMPL-X vertices $V_t^{smpl} \subset \mathcal{S}^c$ into the world space $\mathcal{S}^w$.

To account for the inherent noise and incompleteness of the monocularly reconstructed $\{\mathcal{P}_t^h\}$, we propose a confidence-aware point-to-surface alignment objective. We minimize the weighted residual between the world-space observation $\mathbf{p}_i$ and the transformed parametric model:
\begin{equation}
    \min_{s_t,R_t,t_t}{\sum_{i\in \mathcal{P}_t^n}}w_i\cdot(\min _{v\in V_t^{smpl}}\lVert s_tR_t\textbf{v}+t_t-\mathbf{p}_i\rVert_2)
\end{equation}
where $w_i$ is a multi-modal confidence weight derived from the MVS depth confidence $\mathcal{W}_t$ and the semantic boundary proximity, ensuring that the optimization prioritizes reliable geometric regions (e.g., the torso) over noisy peripherals. 

This process effectively resolves the scale ambiguity inherent in monocular reconstruction and ensures that the subsequent 3DGS avatar is precisely localized within the 3D scene environment, maintaining strict temporal and spatial coherence.

\subsubsection{Canonical 3DGS-based Avatar Generation.} To construct a high-fidelity representation of the human in the target garment, we employ image try-on~\cite{wu2025qwen,hurst2024gpt} and image-to-avatar foundation models~\cite{qiu2025lhm,jiang2025prioravatar,qiu2025anigs} to generate a canonical 3DGS-based avatar. The process initiates with the identification of an optimal keyframe $I_{t^*}$ that maximizes frontal visibility and minimizes self-occlusion, ensuring the extraction of complete texture information. Sampling strategy is in the Supplementary Material.

Following the selection of $I_{t^*}$, we leverage image try-on frameworks to synthesize a high-fidelity reference image $\hat{I}_{t^*}$ featuring the target garment. By feeding $\hat{I}_{t^*}$ and its corresponding SMPL-X pose $\{\Theta_{t^*}\}$ into the avatar foundation model, we distill a canonical 3DGS avatar $\mathcal{G}_{c}$ that captures both the personalized identity and the intricate geometry of the new apparel. We define the Gaussian primitives as $\mathcal{G}_c = \{\mathbf{x}_i, \mathbf{\Sigma}_i, \sigma_i, \mathbf{c}_i\}$, where each primitive is anchored to the upsampled surface of the SMPL-X model in a canonical A-pose $\{\Theta_{c}\}$. Here, $\mathbf{x}_i$, $\mathbf{s}_i$, and $\mathbf{r}_i$ denote the 3D position, anisotropic scale, and quaternion rotation, respectively, while $\sigma_i$ and $\mathbf{c}_i$ capture opacity and view-dependent appearance. By distilling the 2D try-on result into an explicit avatar, we establish a geometrically-grounded reference that inherently maintains structural integrity across varying views.

\subsubsection{Deformation and Rendering.}

To animate the canonical avatar $\mathcal{G}_{c}$, we employ Linear Blend Skinning (LBS) driven by the aligned SMPL-X sequence $\{\Theta_t\}$. As illustrated in Fig.~\ref{fig:pipeline2}(b), the canonical clothed avatar is dynamically warped to match the target poses prescribed by the motion sequence. For each Gaussian primitive, skinning weights $w$ are assigned by querying the nearest vertices of the upsampled SMPL-X mesh in the canonical space. To ensure the avatar is correctly situated within the environment, each primitive’s position $\mathbf{x}$ and covariance $\mathbf{\Sigma}$ are first transformed to the posed space and subsequently mapped to the world space $\mathcal{S}^w$ via the similarity transformation $\mathcal{T}$:
\begin{equation}
\mathbf{x} = \mathcal{T} \left( \sum_{j=1}^{J} w_j \mathbf{T}_j \mathbf{x}_c \right), \quad
\mathbf{\Sigma} = R \left( \sum_{j=1}^{J} w_j \mathbf{R'}_j \mathbf{\Sigma}_c \mathbf{R'}_j^\top \right) R^\top,
\end{equation}
where $\mathbf{T}_j$ and $\mathbf{R'}_j$ denote the rigid bone transformations, and $R$ represents the rotation component of $\mathcal{T}$. This deformation pipeline yields a posed avatar $\mathcal{G}_{p}$ that preserves fine-grained garment details while maintaining strict structural alignment with the background context $\mathcal{P}^b$. 

The reconstructed 4D Try-on Proxy serves as a unified foundation for video synthesis under both input and novel camera trajectories. To synthesize the structural guidance, we adopt a hybrid rendering strategy: the dressed human avatar $\mathcal{G}_p$ is rendered via a differentiable 3DGS rasterizer~\cite{kerbl20233d}, while the background $\mathcal{P}^b$ is processed using a point-based renderer. The detail is in the Supplementary Material.

In summary, by integrating high-fidelity 3DGS-based avatars with persistent background point cloud, our approach yields a geometrically and temporally coherent 4D Try-on Proxy. This representation not only enables high-fidelity rendering across arbitrary camera trajectories but also supports explicit content manipulation within the dynamic 4D scene. By providing versatile spatio-temporal guidance, the proxy effectively bridges the gap between 4D geometric modeling and the subsequent Proxy-Anchored Video Generation stage.

\subsection{Proxy-Anchored Try-on Video Generation}
\label{subsection: Proxy-Anchored Try-on Video Generation}

To synthesize photorealistic try-on videos from the 4D proxy, we present a Video Diffusion Transformer (DiT) anchored by rendered priors. While generic I2V pre-train models often suffer from structural distortion and identity drift under radical motion, our framework circumvents these pitfalls by leveraging pixel-aligned structural metadata to rigorously constrain the generative process. This design ensures physically plausible garment deformations that strictly adhere to the target trajectory $\mathcal{T}$. As shown in Fig.~\ref{fig:pipeline}, TryOnCrafter employs a three-tiered conditioning hierarchy: the Rendered Priors provide foundational spatio-temporal grounding, while the Cross-view Reference Adapter (CRA) and Multi-Modal Semantic Conditioning supply auxiliary appearance and semantic cues to ensure identity fidelity and scene harmony.
\subsubsection{Rendered Prior \& Reference Frame Injection.} 

To enforce strict adherence to the target camera trajectory $\mathcal{T}$, we leverage the rendered video $\mathcal{V}_{render}$ derived from our 4D try-on proxy as a fundamental structural prior. Unlike implicit motion modeling, $\mathcal{V}_{render}$ serves as a pixel-aligned spatio-temporal constraint that explicitly encodes global geometric structures, articulated body poses, and high-fidelity appearance, all precisely synchronized with the prescribed camera trajectory $\mathcal{T}$. By integrating this content-rich rendered prior, the generative model gains a dense structural and texture anchor, ensuring that synthesized garment deformations and character orientations remain geometrically consistent and texture-persistent within the underlying 4D space across all frames. Additionally, to stabilize the global appearance and prevent silhouette drift during the early denoising stages, we inject the high-fidelity reference frame $\hat{I}_{t^*}$ as the first frame of rendered video. 

Technically, the augmented rendered video is processed through the Wan Encoder~\cite{wan2025wan} and integrated with a downsampled binary mask to highlight valid geometric regions. During the patchify stage, these encoded features are concatenated along the channel dimension with the latent noise.

\subsubsection{Cross-view Reference Adapter (CRA).}
While the rendered prior provides global spatio-temporal guidance, the Cross-view Reference Adapter (CRA) is designed to recover fine-grained identity and background details from the source information domain. As illustrated in Fig.~\ref{fig:pipeline2}(c), CRA facilitates a synergistic interaction between the main DiT blocks and the reference branch. For a given DiT feature $\mathbf{F}_i$ and its corresponding reference feature $\mathbf{F}_l$, CRA employs a weight-sharing attention mechanism to bridge the two domains.

To ensure feature alignment while preserving the pre-trained generative prior, CRA shares the key ($W_K$) and value ($W_V$) projection weights with the DiT backbone, but utilizes independent query ($W_Q'$) and output ($W_O'$) weights. The reference residual $\mathbf{O}_l'$ is computed as:

\begin{equation}
\mathbf{O}_l' = \mathrm{Softmax}\left(\frac{(\mathrm{sg}(\mathbf{F}_{i}) W_Q')(\mathbf{F}_{l} W_K)^\top}{\sqrt{d}}\right) (\mathbf{F}_{l} W_V) \cdot W_O'.
\end{equation}

A stop-gradient operator $\mathrm{sg}(\cdot)$ is applied to $\mathbf{F}_{i}$ to maintain training stability. This reference signal is then integrated into the main branch via an additive residual connection:

\begin{equation}
\mathbf{F}_{i}' = \mathrm{Attn}(\mathbf{F}_i, Q_i, K_i, V_i, O_i) + \mathbf{O}_l',
\end{equation}

where $\mathrm{Attn}(\cdot)$ denote the standard attention operation. By complementing the structural guidance of the 4D proxy with these fine-grained source features, the CRA module ensures high-fidelity appearance preservation across unobserved viewpoints with minimal computational overhead.

\begin{figure*}[t]
    \centering
    \setlength{\belowcaptionskip}{-0.5cm}
\includegraphics[width=1.0\linewidth]{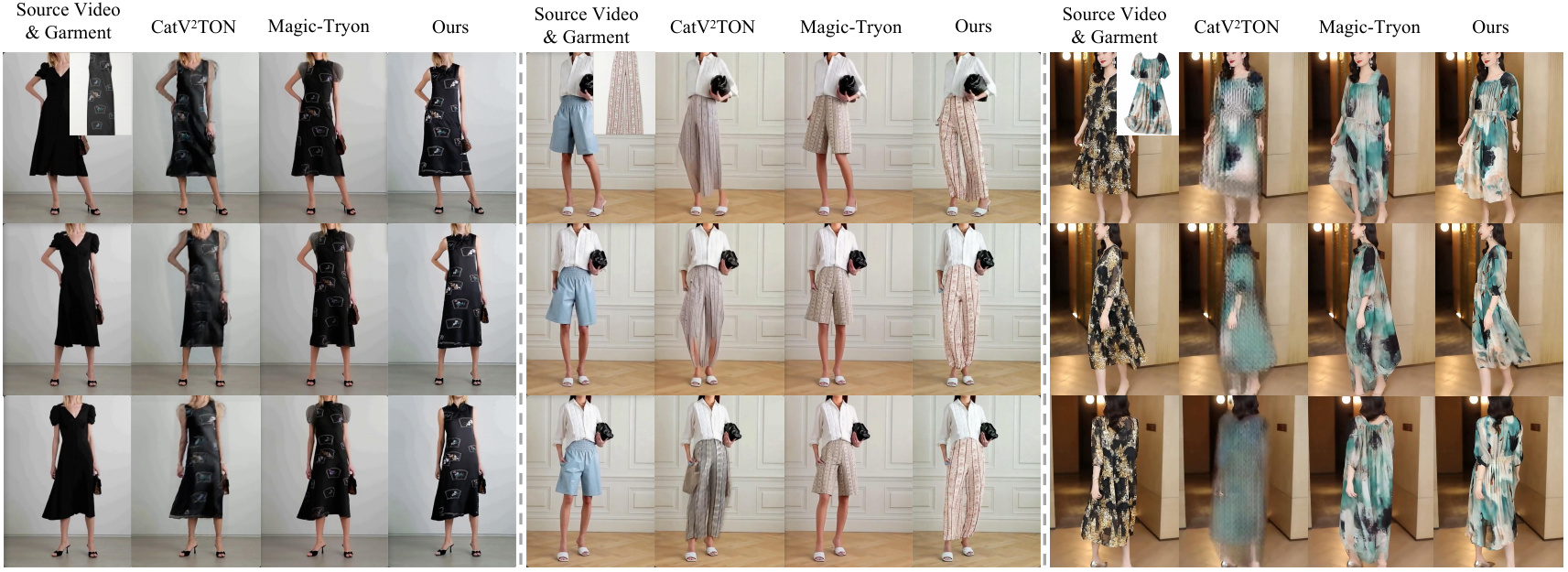}
  \caption{Qualitative comparison on the video try-on benchmark. 
The left two columns show results on ViViD-S\cite{fang2024vivid}, and the right column shows results on the in-the-wild test set. }
  \label{fig:comprison1}
\end{figure*}

\subsubsection{Multi-Modal Semantic Conditioning.}

To complement explicit geometric priors, we incorporate a multi-modal semantic mechanism to ensure global consistency, particularly in occluded or sparsely reconstructed regions. Specifically, we leverage a CLIP image encoder to extract garment features $\mathbf{f}_{gar}$ (texture and material cues) and a UmT5 text encoder for high-level attributes $\mathbf{f}_{text}$. These tokens are concatenated and injected into DiT blocks via cross-attention. This semantic grounding effectively prevents drift in unobserved viewpoints and harmonizes the synthesized apparel with the scene context during long-horizon generation.

\section{Experiments}

\begin{figure*}[t]
    \centering
    \setlength{\belowcaptionskip}{-0.5cm}
\includegraphics[width=1.0\linewidth]{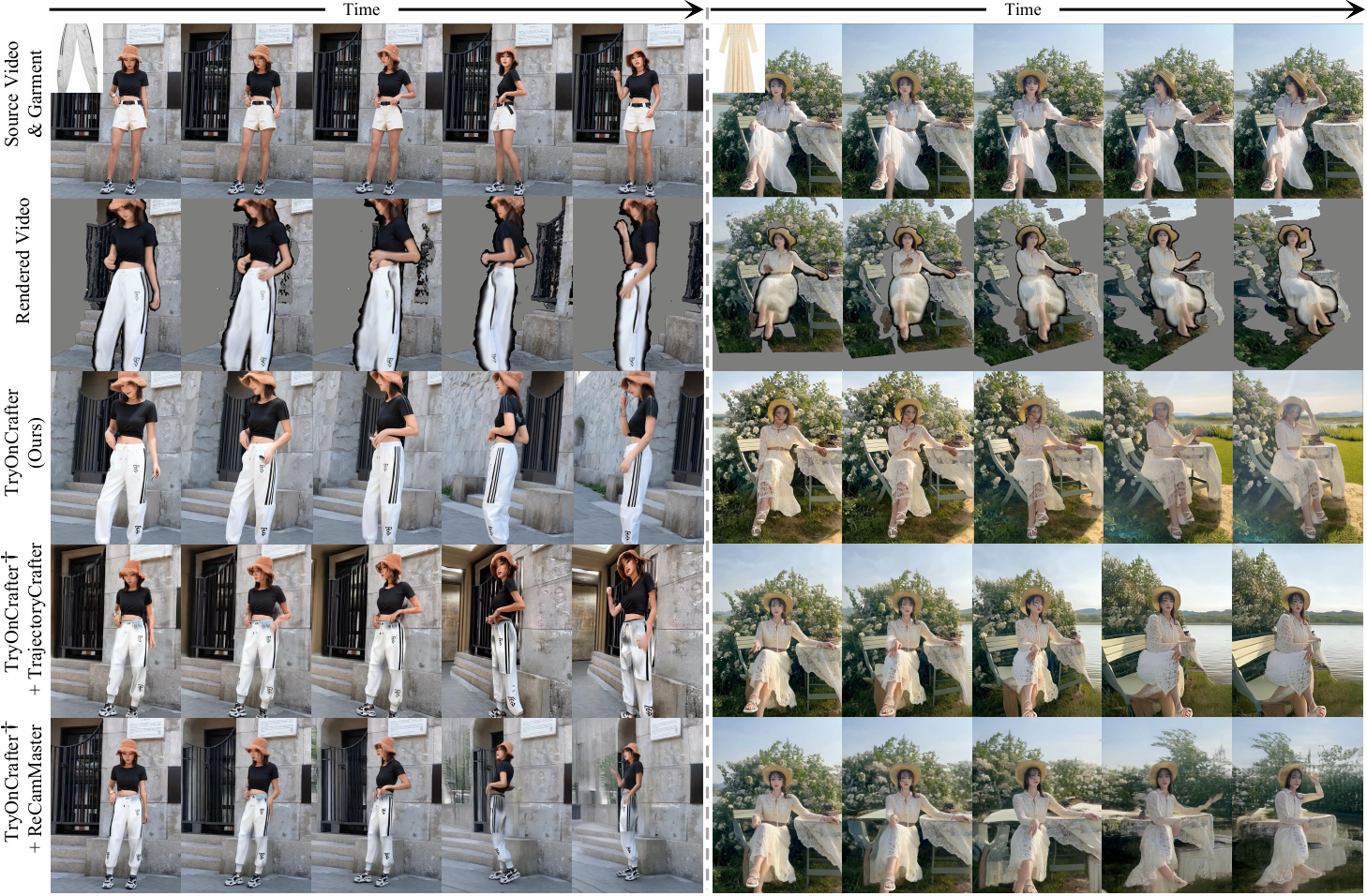}
  \caption{Qualitative comparison results on CaM-VVTBench. \textbf{Left} trajectory: \textit{orbit right} and \textit{zoom in}, \textbf{Right} trajectory: \textit{orbit left}. $^\dagger$ denotes restricted to input trajectories.}
  \label{fig:comprison2}
\end{figure*}
\subsection{Implementation Details}

Our TryonCrafter is built upon the pretrained Wan2.1-I2V-14B~\cite{wan2025wan} foundation, featuring a progressive two-stage training strategy. All experiments are conducted on 16 NVIDIA A100 (80GB) GPUs. For inference, we employ 20 denoising steps and a CFG scale of 6.0.
Detailed training strategies and parameter settings are provided in the \textbf{Supplementary Material}.

\subsection{Dataset and Metrics}

\begin{figure*}[t]
    \centering
    \setlength{\belowcaptionskip}{-0.5cm}
\includegraphics[width=1.0\linewidth]{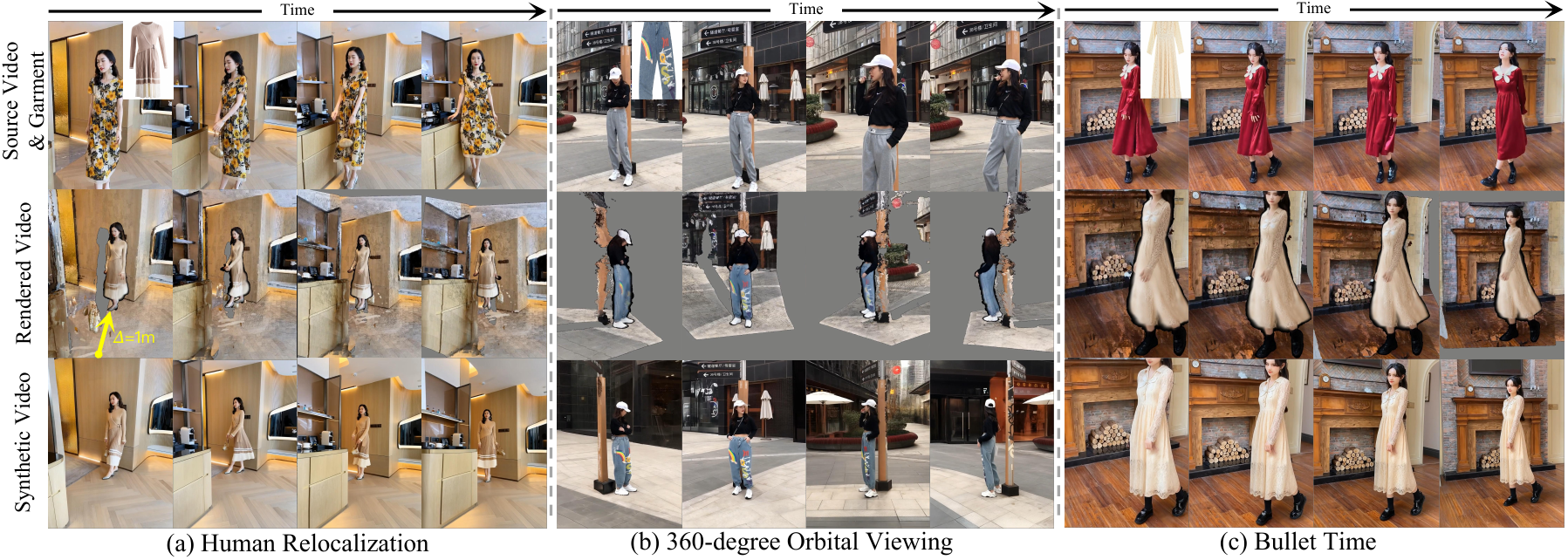}
  \caption{Versatile Applications of TryOnCrafter. Leveraging the decoupled 4D proxy, our model enables controllable synthesis: (a) Human Relocalization: Translating the motion sequence within $\mathcal{S}^w$ while maintaining scene-geometry consistency. (b) 360-degree Orbital Viewing: Synthesizing full orbital trajectories with robust structural integrity in unobserved viewpoints. (c) Bullet Time: Rendering a frozen temporal moment from a moving camera by anchoring the generation to a single pose.}
  \label{fig:application}
\end{figure*}

\noindent\textbf{Video Virtual Try-On Benchmark.}
For VVT evaluation, we adopt the ViViD\cite{fang2024vivid} benchmark, using its official training split (7,759 samples) and evaluating on the ViViD-S test set (180 samples). Following prior works~\cite{li2025magictryon,zuo2025dreamvvt,fang2024vivid}, we adopt VFID with I3D~\cite{carreira2017quo} ($\text{VFID}_I$) and ResNext~\cite{xie2017aggregated} ($\text{VFID}_R$)  under both paired and unpaired settings. For paired evaluation, we additionally report SSIM and LPIPS to measure structural and perceptual similarity to ground truth.

\noindent\textbf{Camera-controllable Video Virtual Try-On Benchmark.}
For CaM-VVT task, we establish a new benchmark, termed CaM-VVTBench. The training set comprises approximately 60K video clips collected from online sources, spanning diverse subjects, garment categories, and camera trajectories to ensure broad coverage of real-world scenarios. The test set contains 96 carefully curated samples with six predefined camera motion patterns, including \emph{tilt up}, \emph{tilt down}, \emph{zoom in}, \emph{zoom out}, \emph{orbit left}, and \emph{orbit right}. These standardized motion primitives enable controlled and reproducible evaluation, forming a dedicated benchmark.
For the evaluation metrics, we use VBench~\cite{huang2024vbench}, which includes Subject Consistency, Imaging Quality, Background Consistency, Aesthetic Quality,  Motion Smoothness, Temporal Flickering,  Overall Consistency scores and Overall Score.

\subsection{Evaluation on Video Virtual Try-On Benchmark}
\noindent\textbf{Baselines.}
We compare our method with state-of-the-art image/video virtual try-on approaches, including StableVITON~\cite{kim2024stableviton}, OOTDiffusion~\cite{xu2025ootdiffusion}, IDM-VTON~\cite{choi2024improving}, ViViD~\cite{fang2024vivid}, CatV2TON~\cite{chong2025catv2ton}, Magic-Tryon~\cite{li2025magictryon}, and DreamVVT~\cite{zuo2025dreamvvt}.
These methods cover diverse paradigms, including GAN-based frameworks, diffusion-based try-on models, and video-aware architectures with temporal constraints.

\noindent\textbf{Quantitative Comparison.} As reported in Table~\ref{table:comprison1}, our method establishes a new SOTA across both paired and unpaired settings, significantly improving video-level fidelity metrics ${\text{VFID}_I}$ and ${\text{VFID}_R}$. In the challenging unpaired scenario, our approach substantially outperforms all baselines. Similarly, top performance in paired ${\text{VFID}_I^p}$ and ${\text{VFID}_R^p}$ reflects enhanced spatiotemporal consistency and garment-aware alignment. While some methods yield marginally higher SSIM in paired cases, our framework delivers superior perceptual quality (LPIPS) while maintaining competitive structural similarity. Overall, these results substantiate the effectiveness of our framework in enhancing video-level realism and coherence across diverse protocols without compromising visual fidelity.

\noindent\textbf{Qualitative Comparison.} Fig.~\ref{fig:comprison1} demonstrates TryOnCrafter's superiority over SOTA frameworks across diverse scenarios. Our method exhibits higher fidelity in texture transfer and color preservation compared to CatV$^2$TON and Magic-Tryon, precisely maintaining intricate patterns and original vibrancy. Notably, by leveraging geometry-aware rendered priors, TryOnCrafter accurately reconstructs complex silhouettes (e.g., full-length trousers), whereas Magic-Tryon fails to resolve structural geometry, erroneously collapsing long garments into shorts. Furthermore, in unconstrained "in-the-wild" environments, our 4D try-on proxy enables the synthesis of realistic dynamics and natural folds for challenging apparel like loose dresses. Conversely, baselines suffer from significant structural distortion and lack the spatiotemporal coherence required for physically plausible deformations in complex scenes.

\begin{table*}[tb]
\centering
\renewcommand\arraystretch{0.9}
\setlength{\tabcolsep}{1.2pt}
{\footnotesize
\begin{minipage}{0.49\linewidth}
\centering
\caption{Quantitative comparison with state-of-the-art methods on the ViViD\cite{fang2024vivid} dataset. The bold represent the best results. $p$ and $u$ denote the paired setting and unpaired setting.}
\scalebox{0.6}{
\begin{tabular}{lcccccc}
\toprule
Method  & ${\text{VFID}_I^p\downarrow}$          & ${\text{VFID}_R^p\downarrow}$             & ${\text{VFID}_I^u\downarrow}$          & ${\text{VFID}_R^u\downarrow}$    & SSIM${\uparrow}$             & LPIPS${\downarrow}$       \\
\midrule
StableVITON  & 34.2446         & 0.7735          & 36.8985          & 0.9064          & 0.8019          & 0.1338          \\
OOTDiffusion & 29.5253         & 3.9372          & 35.3170          & 5.7078          & 0.8087          & 0.1232          \\
IDM-VTON     & 20.0812         & 0.3674          & 25.4972          & 0.7167          & 0.8227          & 0.1163          \\
\midrule
ViViD        & 17.2924         & 0.6209          & 21.8032          & 0.8212          & 0.8029          & 0.1221          \\
CatV2TON     & 13.5962         & 0.2963          & 19.5131          & 0.5283          & 0.8727          & 0.0639          \\
Magic-Tryon   & 12.1988         & 0.2346          & 17.5710          & 0.5073          & \textbf{0.8841} & 0.0815          \\
DreamVVT     & 11.0180         & 0.2549          & 16.9468          & 0.4285          & 0.8737          & 0.0619          \\
Ours         & \textbf{9.6085} & \textbf{0.1817} & \textbf{10.7563} & \textbf{0.2088} & 0.8675          & \textbf{0.0567} \\
\bottomrule
\label{table:comprison1}
\end{tabular}
}
\end{minipage}
\hfill
\begin{minipage}{0.49\linewidth}
\centering
\caption{Ablation study of different try-on components in TryOnCrafter on the ViViD dataset. The bold and underlined entries represent the best and second-best results, respectively. }
\scalebox{0.6}{
\begin{tabular}{lcccccc}
\toprule
Method  & ${\text{VFID}_I^p\downarrow}$          & ${\text{VFID}_R^p\downarrow}$             & ${\text{VFID}_I^u\downarrow}$          & ${\text{VFID}_R^u\downarrow}$    & SSIM${\uparrow}$             & LPIPS${\downarrow}$       \\
\midrule
w/o $\hat{I}_{t^*}$               & 11.8614         & 0.2382          & \underline{18.1947}          & \underline{0.2996}          & 0.8526          & 0.0669          \\
w/o mask                   & 9.9393          & \underline{0.1927}         & 18.5710          & 0.4212          & 0.8633          & 0.0582          \\
w/o $\textbf{f}_{text}$                & 18.7144         & 0.3500          & 18.5158          & 0.3707          & 0.8189          & 0.1319 \\
w/o $\textbf{f}_{gar}$ & \textbf{9.1740} & 0.2257          & 18.4331          & 0.3487 & \textbf{0.8680} & \textbf{0.0562}         \\
w/o $\mathcal{V}_{render}$ & 16.4608 & 0.3937 & 20.1280 & 0.4159 & 0.8346 & 0.0925 \\
full (Ours)                       & \underline{9.6085} & \textbf{0.1817} & \textbf{10.7563} & \textbf{0.2088} & \underline{0.8675} & \underline{0.0567} \\
\bottomrule
\label{tab:ablation}
\end{tabular}
}
\end{minipage}

}

\centering
\setlength{\abovecaptionskip}{0.5cm}
\caption{Quantitative comparison with two-stage methods on the CaM-VVTBench. The bold represent the best results. $^\dagger$ denotes restricted to input trajectories.}
\renewcommand\arraystretch{0.8}
\setlength{\tabcolsep}{1.2pt}
\resizebox{1.0\linewidth}{!}{
\begin{tabular}{lcccccccc}
\toprule

Method & \makecell[c]{Overall\\Score${\uparrow}$} & \makecell[c]{Subject\\Consis.${\uparrow}$} & \makecell[c]{Imaging\\Quality${\uparrow}$} & \makecell[c]{Background\\Consis.${\uparrow}$} & \makecell[c]{Aesthetic\\Quality${\uparrow}$}  & \makecell[c]{Motion\\Smooth.${\uparrow}$} & \makecell[c]{Temporal\\Flicker.${\uparrow}$} & \makecell[c]{Overall\\Consis.${\uparrow}$}\\
\midrule
Magic-Tryon+TrajectoryCrafter  & 71.64     & 87.76      & 68.27           & 90.76        & 49.89    & 97.88   & 94.53   & 5.15            \\
Magic-Tryon+ReCaMaster         & 61.92    & 87.93       & 66.69     & 91.82         & 46.76           & 98.39  & 95.50   & 8.29       \\
TryOnCrafter$^\dagger$+TrajectoryCrafter & 71.95    &   88.52    & 68.93    & 91.18        & 51.29    & 97.93    & 94.60      & 7.53             \\
TryOnCrafter$^\dagger$+ReCaMaster        & 74.32  & 88.90  & 67.70         & 91.72            & 47.64       & \textbf{98.45}   & \textbf{95.54} & 8.35       \\
TryOnCrafter (Ours)             & \textbf{75.47}      & \textbf{92.71}  & \textbf{74.28}  & \textbf{92.05}  & \textbf{51.44}        & 98.32     & 94.90  & \textbf{9.61}    \\
\bottomrule
\end{tabular}
}
\label{tab:comprison2}
\end{table*}

\subsection{Evaluation on CaM-VVTBench}
\noindent\textbf{Baselines.}
Given the absence of end-to-end models for the CaM-VVT task, we establish rigorous benchmarks by coupling open-source state-of-the-art (SOTA) VVT methods with representative camera control frameworks. Specifically, we employ Magic-Tryon~\cite{li2025magictryon} as the VVT baseline.
For camera steering, we integrate two distinct paradigms: TrajectoryCrafter~\cite{yu2025trajectorycrafter}, which leverages explicit 3D point cloud reconstruction and geometric priors, and ReCaMaster~\cite{bai2025recammaster}, which utilizes implicit parameter injection.
Furthermore, to ensure a comprehensive and fair comparison, we evaluate combinations of these control methods with TryOnCrafter$^\dagger$ ( $^\dagger$ denotes \textbf{restricted to input trajectories}).

\noindent\textbf{Qualitative Comparison.} As illustrated in Fig. \ref{fig:comprison2}, while two-stage baselines achieve plausible initial synthesis by leveraging the try-on priors of TryOnCrafter$^\dagger$, their performance degrades significantly as camera movements escalate in intensity and perspective complexity. Specifically, TrajectoryCrafter frequently exhibits structural breakdown and limb distortions (e.g., subjects' arms) during aggressive rotations due to the lack of explicit human priors and its reliance on fragmented 3D point clouds. In contrast, our unified framework provide the human avatar in Renderable 4D Try-on Proxy as a continuous geometric foundation, significantly outperforming the point-cloud baseline in preserving limb integrity and background coherence. Furthermore, ReCaMaster suffers from a profound deficit in structural comprehension; the absence of explicit geometric constraints yields severe warping across both the human subject and the complex background manifold (e.g., distorted bushes and bench). Most crucially, while baselines fail to maintain texture consistency or generate physically plausible clothing motion under violent maneuvers, our TryOnCrafter effectively achieves viewpoint-agnostic synthesis of fine-grained textures from unobserved angles. By synthesizing realistic, trajectory-aligned cloth deformations, our Proxy-Anchored Video DiT demonstrates superior structural robustness and high-fidelity appearance across unconstrained trajectories.

\noindent\textbf{Quantitative Comparison.}
As shown in Table~\ref{tab:comprison2}, TryOnCrafter establishes a new SOTA on CaM-VVTBench, significantly outperforming two-stage baselines. Our model leads substantially in Overall Score, Subject Consistency, and Imaging Quality, underscoring the efficacy of the Proxy-Anchored DiT in mitigating error accumulation and preserving fine-grained textures. While ReCaMaster-based variants show marginal gains in Motion Smoothness, qualitative results reveal this often stems from excessive blurring and identity loss during aggressive maneuvers. Conversely, TryOnCrafter achieves a superior equilibrium between temporal stability and high-fidelity synthesis, evidenced by top-tier Aesthetic Quality and Background Consistency. These metrics validate the Renderable 4D Proxy as a robust structural foundation for physically plausible, 3D-aware virtual try-on.

\begin{figure*}[t]
    \centering
    \setlength{\belowcaptionskip}{-0.5cm}
\includegraphics[width=1.0\linewidth]{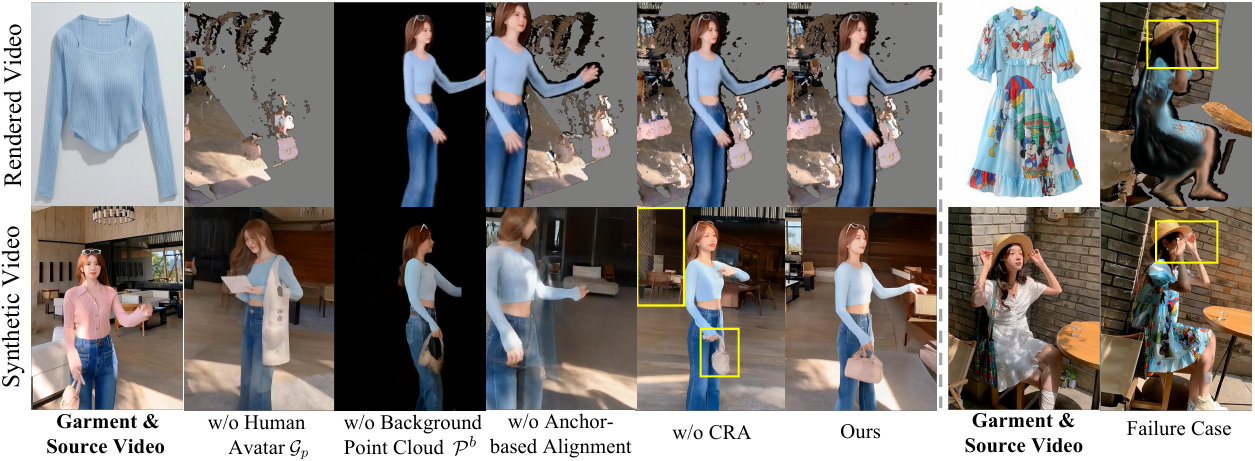}
  \caption{\textbf{Left}: Ablation studies of main component of TryOnCrafter. \textbf{Right}: Failure case caused by perspective-induced parallax and inaccuracies of SMPL-X estimation.}
  \label{fig:ablation_and_failurecase}
\end{figure*}

\subsection{Ablation Study}

\noindent\textbf{Ablation on 4D Try-on Proxy.} We systematically evaluate the core components of the 4D Try-on Proxy—the clothed avatar $\mathcal{G}_p$, the background point cloud $\mathcal{P}^b$, and the Anchor-based Alignment—using the CaM-VVTBench. As illustrated in Fig.~\ref{fig:ablation_and_failurecase}, omitting $\mathcal{G}_p$ deprives the model of essential pose and texture priors, leading to stochastic motion and severe structural artifacts. The absence of $\mathcal{P}^b$ restricts the generative scope solely to the human subject, failing to capture the environmental context and ambient lighting. Furthermore, removing the Anchor-based Alignment disrupts spatio-temporal synchronization, resulting in misaligned poses and discontinuous trajectories.

\noindent\textbf{Ablation on DiT Conditioning Techniques.} We further investigate the conditioning mechanisms in Proxy-Anchored Video DiT, comprising rendered priors, the CRA module, and multimodal semantic cues. Quantitative results in Table~\ref{tab:ablation} confirm the necessity of this integration; notably, the significant performance decline in the w/o Rendered Video setting underscores that explicit geometric anchoring is indispensable for maintaining structural integrity and visual fidelity. Regarding specific components, while removing garment semantics $\mathbf{f}_{gar}$ yields marginal gains in simple paired settings, it causes severe degradation in unpaired scenarios, highlighting the importance of semantic priors for generalization. Furthermore, omitting the CRA module—responsible for source identity and articulated cues—drastically reduces cross-view geometric consistency and ID preservation (Fig.~\ref{fig:ablation_and_failurecase}). These findings substantiate that the synergy between our explicit 4D proxy and multi-modal conditioning is vital for high-fidelity, camera-controllable try-on.
\section{Applications}
The explicit 4D Try-on Proxy enables the decoupling of human subjects, background, and camera trajectories within a unified world space $\mathcal{S}^w$. Beyond standard synthesis, we explore three generative applications that demonstrate the framework's versatility.

\noindent\textbf{Human Relocalization.} By integrating the subject and background into $\mathcal{S}^w$, we can perform arbitrary spatial edits on the motion sequence. As shown in Fig.~\ref{fig:application}(a), our geometry-grounded approach facilitates Human Relocalization, ensuring consistent occlusion and metric-scale harmony between the edited subject trajectory and the environment—a task where traditional 2D-based methods often suffer from perspective distortion.

\noindent\textbf{360-degree Orbital Viewing.} While existing frameworks~\cite{yu2025trajectorycrafter,bai2025recammaster} are often limited to narrow-angle synthesis (e.g., $\pm60^{\circ}$), TryOnCrafter achieves high-fidelity 360-degree orbital viewing generation. By leveraging the 4D try-on proxy as a persistent geometric anchor, our model maintains structural stability and texture consistency even in unobserved viewpoints (Fig.~\ref{fig:application}(b)), effectively resolving the challenges of extreme disocclusion.

\noindent\textbf{Bullet Time Virtual Try-on.} We extend our framework to Bullet Time effects, where a moving camera renders a frozen temporal instant. This is achieved by anchoring the generation to a specific SMPL-X pose $\Theta_{t^*}$ and its corresponding canonical frame $I_{t^*}$ across the sequence. As illustrated in Fig.~\ref{fig:application}(c), the resulting videos exhibit smooth, view-consistent transitions around a single pose, offering a novel immersive try-on experience.
\section{Limitation}

Despite its strong performance, TryOnCrafter possesses certain limitations. Specifically, our framework relies on a parametric 4D Try-on Proxy for geometric guidance; however, extreme viewpoint transitions often introduce significant parallax and ambiguity, challenging the structural alignment between the subject and the background. As shown in Fig.~\ref{fig:ablation_and_failurecase}, such spatial shifts can exacerbate estimation inaccuracies in the SMPL-X model, occasionally manifesting as misaligned hand poses or subtle geometric inconsistencies during radical articulated motions. Furthermore, the iterative denoising process of our DiT-based framework incurs high inference costs, hindering real-time interaction for trajectory edits.
\section{Conclusion}

We pioneer Camera-controllable Video Virtual Try-on (CaM-VVT) to overcome the trajectory dependency of existing paradigms. Our framework, TryOnCrafter, integrates two synergistic components: a Renderable 4D Try-on Proxy for explicit subject-environment decoupling, and a Proxy-anchored DiT that leverages a multi-tiered conditioning hierarchy for high-fidelity synthesis. By anchoring the generative process to the 4D proxy, TryOnCrafter ensures physically plausible garment deformations and temporal coherence across unconstrained camera movements. Extensive evaluations on CaM-VVTBench demonstrate that our method significantly outperforms SOTA baselines. Furthermore, the 4D representation’s versatility enables diverse applications like human relocalization and 360-degree orbital viewing, paving a new way for interactive digital fashion.

\bibliographystyle{splncs04}
\bibliography{main}

\clearpage



\setcounter{section}{0}
\setcounter{subsection}{0}
\setcounter{figure}{0}
\setcounter{table}{0}
\setcounter{equation}{0}

\renewcommand{\thesection}{S\arabic{section}}
\renewcommand{\thesubsection}{S\arabic{section}.\arabic{subsection}}
\renewcommand{\thefigure}{S\arabic{figure}}
\renewcommand{\thetable}{S\arabic{table}}
\renewcommand{\theequation}{S\arabic{equation}}

\section*{Supplementary Materials}
\addcontentsline{toc}{section}{Supplementary Materials}

\section{Overview}
In the supplementary materials, we provide additional implementation details, analyses, and qualitative results for TryOnCrafter. Specifically, we include:
\begin{itemize}
    \item Training scheme and dataset construction in Sec.~\ref{sec:supp_training_dataset};
    \item Keyframe sampling strategy in Sec.~\ref{sec:supp_keyframe_sampling};
    \item Hybrid rendering strategy in Sec.~\ref{sec:supp_hybrid_rendering};
    \item Benchmark scale and baseline comparison in Sec.~\ref{sec:supp_benchmark};
    \item Efficiency and reusability analysis in Sec.~\ref{sec:supp_efficiency};
    \item Sensitivity and robustness analysis in Sec.~\ref{sec:supp_robustness};
    \item More comparisons with SOTA methods in Sec.~\ref{sec:supp_sota_comparison};
    \item More qualitative results in Sec.~\ref{sec:supp_more_results}.
\end{itemize}

\section{Training Scheme and Dataset Construction}
\label{sec:supp_training_dataset}

\subsubsection{Training Scheme.}
As shown in Table~\ref{tab:supp_training_recipe}, we adopt a progressive two-stage training strategy to achieve high-fidelity and spatially consistent multi-view virtual try-on.

\textbf{Stage 1: Base VVT Pre-training.}
The model is initialized from the pre-trained Wan2.1-I2V foundation model and trained under standard monocular video virtual try-on settings. This stage aims to establish a robust foundation for high-fidelity texture transfer and spatiotemporal coherence. To maintain training stability while preserving visual details, we use a coarse-to-fine resolution schedule, progressively increasing the input resolution to help the model capture intricate garment textures and fine-grained patterns.

\textbf{Stage 2: CaM-VVT Fine-tuning.}
In this stage, we introduce camera control and continue training at full resolution to enforce cross-view spatial consistency. This phase encourages the model to learn view-aware garment deformation and geometry-consistent rendering. As a result, the model can move beyond static-viewpoint synthesis and generate plausible human appearances under dynamic and unconstrained camera trajectories.

\begin{table*}[tb]
\centering
\caption{Training strategies and parameter settings.}
\label{tab:supp_training_recipe}
\setlength{\tabcolsep}{3.5pt}
\renewcommand{\arraystretch}{0.95}
\small
\resizebox{0.82\textwidth}{!}{
\begin{tabular}{lccc}
\toprule
 & Stage 1 (Low) & Stage 1 (High) & Stage 2 \\
\midrule
Resolution & 256--512 & 512--1024 & 512--1024 \\
Video clip length & 81 frames & 49 frames & 49 frames \\
Training steps & 24,000 & 24,000 & 24,000 \\
Learning rate & $1\times10^{-5}$ & $1\times10^{-6}$ & $1\times10^{-6}$ \\
Warm-up steps & 100 & 200 & 200 \\
\midrule
Weight decay & \multicolumn{3}{c}{0.03} \\
Gradient clipping & \multicolumn{3}{c}{0.05, L2 norm} \\
Global batch size & \multicolumn{3}{c}{16} \\
Optimizer & \multicolumn{3}{c}{AdamW, $\beta_1=0.9$, $\beta_2=0.999$} \\
\bottomrule
\end{tabular}
}
\vspace{-0.3cm}
\end{table*}

\subsubsection{Training Dataset Construction.}
The scarcity of synchronized multi-view try-on data poses a major challenge for training CaM-VVT models. To address this issue, we construct different training data for the two stages.

\textbf{Monocular Try-on Dataset for Stage 1.}
For the initial VVT pre-training stage, we use monocular try-on videos from the ViViD benchmark together with high-quality in-the-wild sequences curated for CaM-VVTBench. This data composition improves robustness across different subjects, garments, poses, and environments. For experiments on the ViViD benchmark, we restrict the training data to ViViD to ensure a fair and controlled comparison under the standard evaluation protocol.

\begin{figure*}[t]
\centering
\includegraphics[width=1.0\linewidth]{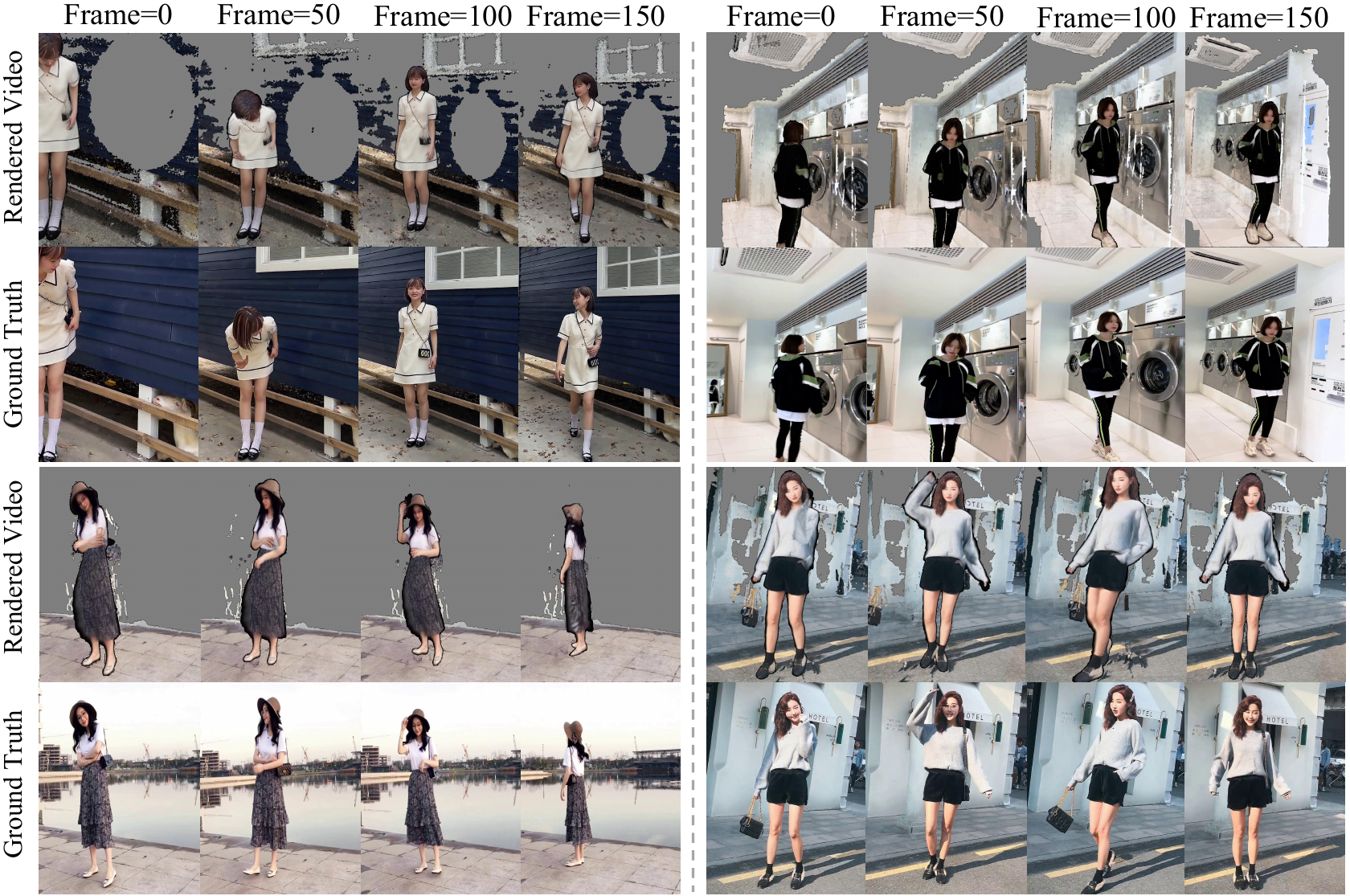}
\caption{Visualized examples of the synthesized multi-view pairs in CaM-VVTBench. Each pair consists of a rendered proxy input and its corresponding photorealistic target.}
\label{fig:supp_dataset}
\end{figure*}

\textbf{Synthetic Multi-view Try-on Dataset for Stage 2.}
Since synchronized multi-view try-on videos are expensive to acquire, we follow recent synthesis paradigms~\cite{yu2025trajectorycrafter,yu2024viewcrafter,wu2025difix3d+} and construct a large-scale synthetic multi-view try-on dataset. Specifically, we transform monocular in-the-wild videos from CaM-VVTBench into multi-view training pairs using a dual-reprojection scheme combined with randomized 2D/3D masking strategies. As illustrated in Fig.~\ref{fig:supp_dataset}, each synthesized pair consists of a rendered proxy view and its corresponding photorealistic target frame. These pairs simulate complex camera movements and provide dense structural supervision for the DiT to learn view-aware and view-invariant representations.

\section{Keyframe Sampling Strategy}
\label{sec:supp_keyframe_sampling}

For each frame $t$, we estimate a torso forward vector $\mathbf{f}_t$ by computing the normalized cross product between the shoulder-to-shoulder vector and the neck-to-pelvis vector derived from the SMPL-X joints. We then select the canonical frame by maximizing the viewing score $s_t$, which measures the alignment between $\mathbf{f}_t$ and the camera optical axis $-\mathbf{z}_c$:
\begin{equation}
t^* = \arg\max_{t} 
\frac{\mathbf{f}_t \cdot (-\mathbf{z}_c)}
{|\mathbf{f}_t| |\mathbf{z}_c|}.
\end{equation}

This strategy favors frames where the human body is well aligned with the camera view, providing a stable canonical reference for subsequent proxy construction and try-on synthesis.

\section{Hybrid Rendering Strategy}
\label{sec:supp_hybrid_rendering}

Given the reconstructed 4D try-on proxy, we can synthesize videos under either the input camera trajectory or novel camera trajectories. We adopt a hybrid rendering strategy: the human avatar $\mathcal{G}^p$ is rendered with the 3D Gaussian Splatting pipeline~\cite{kerbl20233d}, while the background point cloud $\mathcal{P}^b$ is rendered using the PyTorch3D rasterizer. To handle mutual occlusions between the human avatar and the background, we fuse the rendered outputs through depth-buffer comparison:
\begin{equation}
    C(p)=
    \begin{cases}
        C_b(p), 
        & \text{if } D_b(p) < D_h(p), \\
        \displaystyle
        \sum_{i \in \mathcal{N}} c_i \alpha_i 
        \prod_{j=1}^{i-1} (1-\alpha_j), 
        & \text{if } D_b(p) \ge D_h(p),
    \end{cases}
\end{equation}
where $D_b$ and $D_h$ denote the rendered depths of the background and human avatar, respectively. This hybrid rendering scheme produces geometrically consistent and temporally coherent video proxies, which serve as spatial-temporal guidance for the subsequent generative refinement stage.

\section{Benchmark Scale and Baseline Comparison}
\label{sec:supp_benchmark}

CaM-VVTBench requires labor-intensive trajectory annotation to ensure accurate camera control across high-resolution indoor and outdoor videos. Therefore, its scale is moderate, but it is comparable to existing video virtual try-on benchmarks, as shown in Table~\ref{tab:supp_benchmark_comparison}. Compared with prior benchmarks, CaM-VVTBench covers both indoor and outdoor scenes and provides camera-trajectory-aware evaluation samples, making it suitable for evaluating camera-controllable video virtual try-on.

\begin{table*}[t]
\centering
\caption{Comparison of video virtual try-on benchmarks.}
\label{tab:supp_benchmark_comparison}
\setlength{\tabcolsep}{3pt}
\renewcommand{\arraystretch}{0.9}
\small
\resizebox{0.92\textwidth}{!}{
\begin{tabular}{lccccc}
\toprule
Benchmark 
& VVT 
& ViViD-S 
& VVT-Interact 
& Wild-TryOnBench 
& \textbf{CaM-VVTBench} \\
\midrule
Scene type 
& Indoor 
& Indoor 
& Indoor 
& Indoor \& outdoor 
& \textbf{Indoor \& outdoor} \\
Resolution 
& 256$\times$192 
& 512$\times$384 
& 576$\times$768 
& 720P 
& \textbf{480P \& 720P} \\
Test set size 
& 130 
& 180 
& 132 
& 81 
& \textbf{96} \\
Camera trajectory control
& No
& No
& No
& No
& \textbf{Yes} \\
\bottomrule
\end{tabular}
}
\end{table*}

\section{Efficiency and Reusability Analysis}
\label{sec:supp_efficiency}

\begin{figure*}[t]
    \centering
    \includegraphics[width=0.95\linewidth]{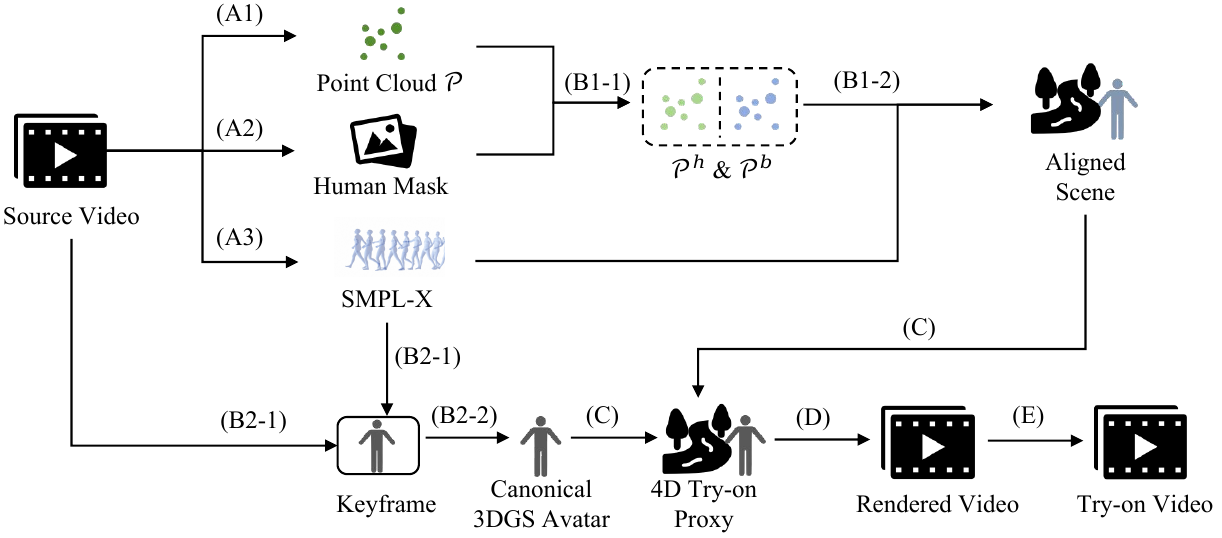}
    \caption{Runtime profiling pipeline of TryOnCrafter. Several preprocessing stages are independent or decoupled, making them naturally parallelizable.}
    \label{fig:supp_pipeline_runtime}
\end{figure*}

We provide detailed runtime and memory profiling on a single A100 GPU with 80GB memory. As shown in Fig.~\ref{fig:supp_pipeline_runtime} and Table~\ref{tab:supp_runtime}, the full proxy construction and rendering pipeline takes 38.9s with 59.1GB peak memory. In contrast, video generation takes 1360.0s with 68.6GB peak memory at 720P, indicating that the main computational bottleneck comes from the 14B DiT backbone rather than the proxy construction pipeline.

Many preprocessing stages are naturally parallelizable: point-cloud reconstruction, segmentation, and SMPL-X estimation can be executed independently; foreground-background separation and keyframe sampling are decoupled; avatar deformation and proxy rendering can also be batched across different camera trajectories.

\begin{table*}[t]
\centering
\caption{Component-wise runtime and peak memory usage on a single A100 GPU.}
\label{tab:supp_runtime}
\setlength{\tabcolsep}{5pt}
\renewcommand{\arraystretch}{0.9}
\small
\resizebox{0.72\textwidth}{!}{
\begin{tabular}{lcc}
\toprule
Component & Time (s) & Peak memory (GB) \\
\midrule
(A1) Point-cloud reconstruction & 0.6 & 5.9 \\ 
(A2) SAM2 segmentation & 6.3 & 7.8 \\
(A3) SMPL-X estimation & 4.2 & 0.5 \\
(B1-1) $\mathbf{P}^h/\mathbf{P}^b$ separation & $<$0.1 & 5.9 \\
(B1-2) Anchor alignment & 4.3 & 3.9 \\
(B2-1) Keyframe sampling & $<$0.1 & 0 \\
(B2-2) Image try-on \& avatar generation & 19.1 & 59.1 \\
(C) Avatar deformation & 1.3 & 20.3 \\
(D) Proxy rendering & 14.3 & 20.3 \\
\midrule
(A-D) Proxy total & \textbf{38.9} & \textbf{59.1} \\
(E) Video generation & \textbf{1360.0} & \textbf{68.6} \\
\bottomrule
\end{tabular}
}
\end{table*}

Table~\ref{tab:supp_efficiency_comparison} further compares TryOnCrafter with representative baselines. TryOnCrafter has a generation cost comparable to the Wan2.1-I2V backbone, while being much faster than two-stage baselines. More importantly, the constructed 4D proxy is reusable: once the full proxy is built, changing the camera trajectory only requires re-rendering the proxy, which takes 14.3s with 20.3GB peak memory. This substantially reduces the marginal cost when synthesizing multiple camera trajectories for the same input video and garment.

\begin{table*}[t]
\centering
\caption{Efficiency comparison in terms of runtime / peak memory. Runtime is measured in seconds and memory in GB.}
\label{tab:supp_efficiency_comparison}
\setlength{\tabcolsep}{3.5pt}
\renewcommand{\arraystretch}{0.9}
\small
\resizebox{0.95\textwidth}{!}{
\begin{tabular}{llccccc}
\toprule
Model 
& Resolution 
& Full preprocess 
& Re-trajectory 
& 1 GPU 
& 4 GPUs 
& 8 GPUs \\
\midrule
Wan2.1-I2V-14B
& 480P & / & / & 386 / 48.3 & 117 / 28.4 & 97 / 23.2 \\
Wan2.1-I2V-14B
& 720P & / & / & 1265 / 67.0 & 357 / 38.3 & 208 / 31.3 \\
\midrule
Magic-TryOn + TrajectoryCrafter
& 480P & 40 / 22.7 & 35 / 22.7 & 850 / 66.6 & -- & -- \\
Magic-TryOn + TrajectoryCrafter
& 720P & -- & -- & -- & -- & -- \\
\midrule
Magic-TryOn + ReCaMaster
& 480P & 5 / 0.8 & / & 810 / 48.3 & -- & -- \\
Magic-TryOn + ReCaMaster
& 720P & -- & -- & -- & -- & -- \\
\midrule
\textbf{TryOnCrafter}
& 480P & 39 / 59.1 & 14 / 20.3 & 391 / 49.0 & 142 / 41.5 & 86 / 28.4 \\
\textbf{TryOnCrafter}
& 720P & 39 / 59.1 & 14 / 20.3 & 1360 / 68.6 & 476 / 63.1 & 268 / 39.3 \\
\bottomrule
\end{tabular}
}
\end{table*}

\section{Sensitivity and Robustness Analysis}
\label{sec:supp_robustness}

\begin{figure*}[t]
    \centering
    \includegraphics[width=1.0\linewidth]{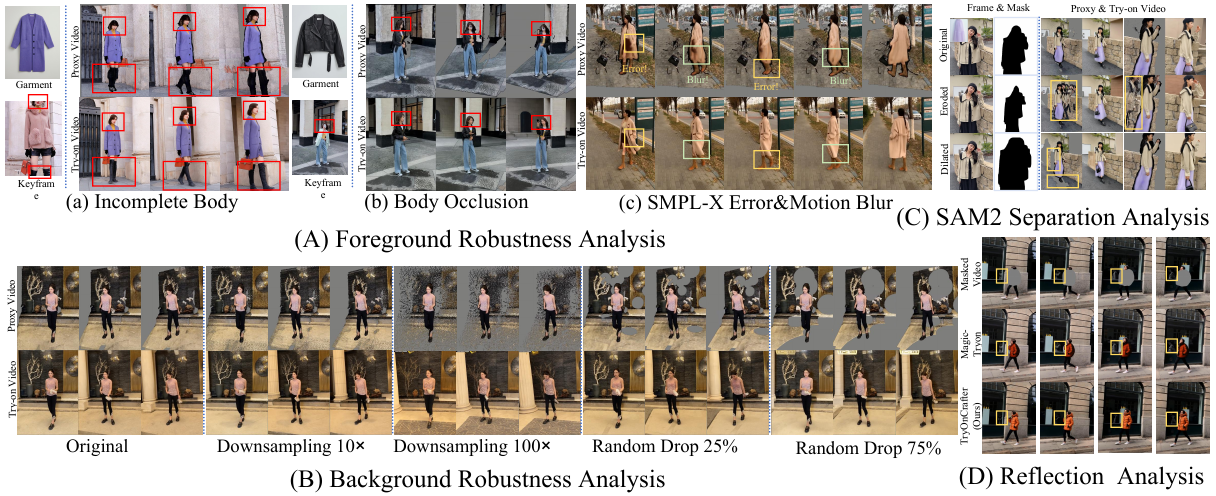}
    \caption{Sensitivity and robustness analysis of TryOnCrafter. We evaluate robustness to foreground errors, background point-cloud degradation, segmentation perturbations, and reflection-related failure cases. Please zoom in for details.}
    \label{fig:supp_robustness_analysis}
\end{figure*}

We further conduct stress tests to analyze the robustness of TryOnCrafter against upstream errors. As shown in Fig.~\ref{fig:supp_robustness_analysis}, TryOnCrafter remains stable under several challenging perturbations.

\textbf{Foreground reconstruction errors.}
We test cases with incomplete body reconstruction, occlusion, SMPL-X estimation errors, and motion blur. Despite imperfect upstream reconstruction, TryOnCrafter still preserves coherent human motion and stable garment structure.

\textbf{Background reconstruction errors.}
We further evaluate robustness under degraded background point clouds, including 10$\times$/100$\times$ point-cloud downsampling and 25\%/75\% random point dropping. TryOnCrafter maintains reasonable background consistency because the proxy is used as coarse geometric guidance rather than a strict pixel-level target.

\textbf{Segmentation errors.}
We also perturb the SAM2 masks with erosion and dilation. The results show that TryOnCrafter is robust to moderate segmentation errors, which is important for in-the-wild videos where segmentation masks are often imperfect.

This robustness mainly comes from two aspects. First, the renderable 4D proxy provides structured geometric guidance without forcing the generator to exactly copy noisy proxy pixels. Second, our training augmentation exposes the DiT to noisy and incomplete proxy conditions through dual reprojection and randomized 2D/3D masking.

\subsubsection{Reflection-related Failure Cases.}
As shown in Fig.~\ref{fig:supp_robustness_analysis} (D), unchanged reflections may appear in mirror-like scenes. This issue is mainly caused by residual reflection cues in the masked source video. Since the model may still observe partial reflection evidence during generation, it can preserve the original reflection rather than synthesizing a physically correct one under the new try-on condition.

\section{More Comparisons with SOTA Methods}
\label{sec:supp_sota_comparison}

\begin{figure*}[t]
    \centering
    \includegraphics[width=1.0\linewidth]{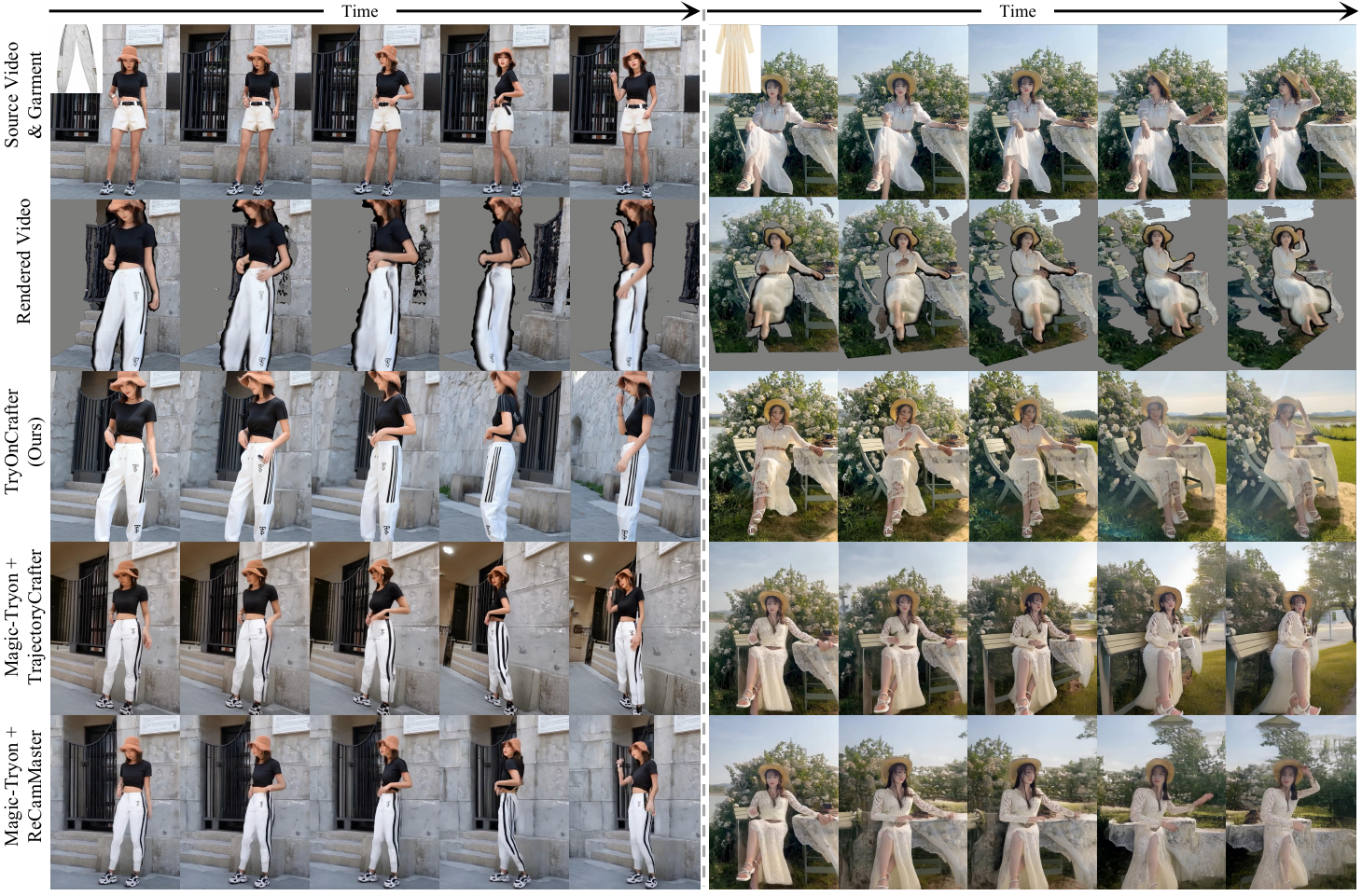}
    \caption{Qualitative comparison with Magic-TryOn-based methods.}
    \label{fig:supp_comparison}
\end{figure*}

Fig.~\ref{fig:supp_comparison} presents additional qualitative comparisons with Magic-TryOn-based methods. The results further demonstrate that TryOnCrafter produces more faithful garment details, more stable identity preservation, and more plausible visual quality under challenging camera trajectories.

\section{More Results of TryOnCrafter}
\label{sec:supp_more_results}

\begin{figure*}[t]
    \centering
    \includegraphics[width=1.0\linewidth]{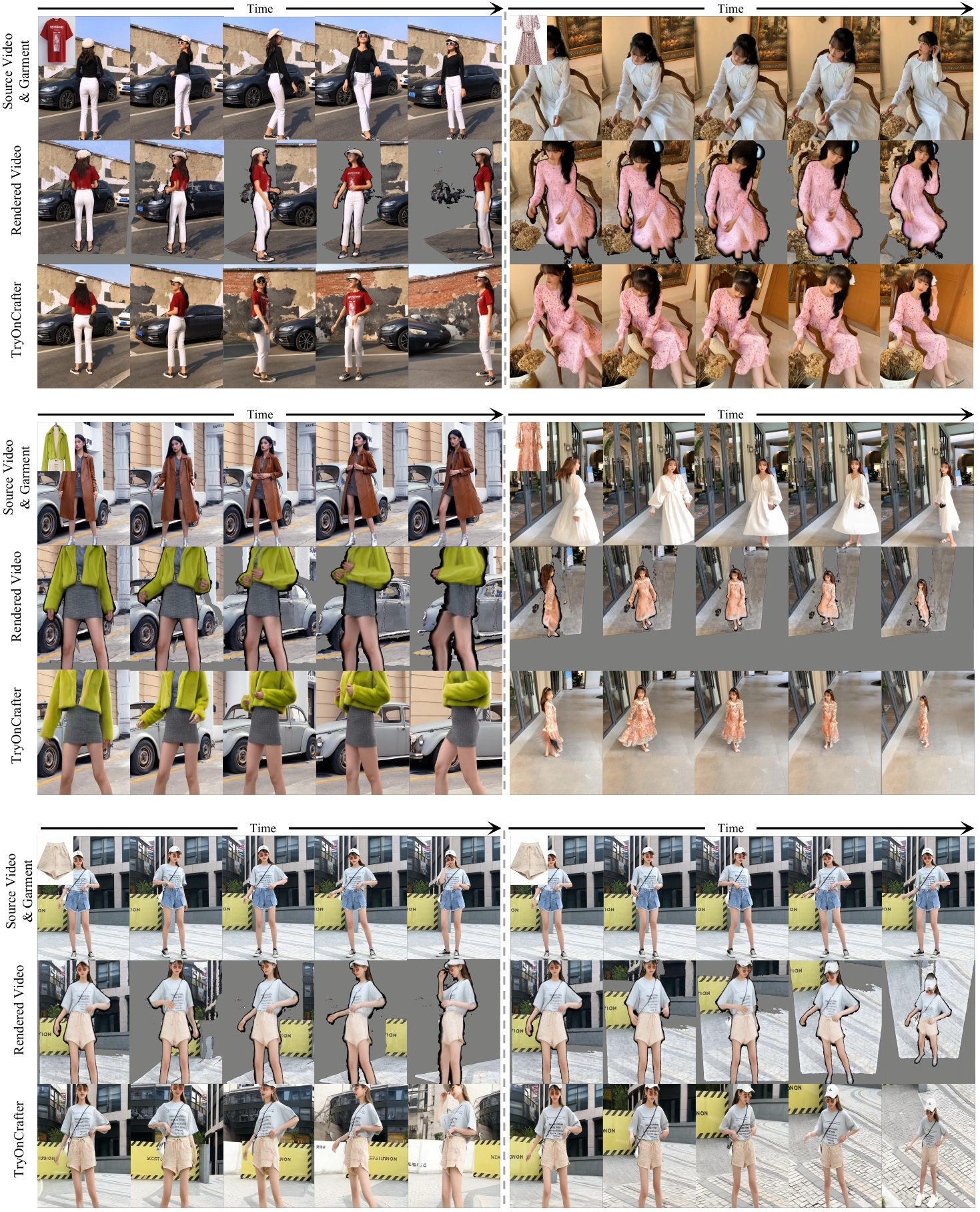}
    \caption{More synthesized results of TryOnCrafter.}
    \label{fig:supp_results1}
\end{figure*}

More synthesized results of TryOnCrafter are shown in Fig.~\ref{fig:supp_results1}. Please refer to our demo video for additional dynamic results.

\end{document}